\documentclass{article}
\usepackage{spconf,amsmath,graphicx}
\usepackage{bm,amssymb,array}
\usepackage{fancyhdr}

\title{A MULTICHANNEL CONVOLUTIONAL NEURAL NETWORK\\
FOR CROSS-LANGUAGE DIALOG STATE TRACKING}
\name{Hongjie Shi, Takashi Ushio, Mitsuru Endo, Katsuyoshi Yamagami and Noriaki Horii}
\address{Interactive AI Research Group, Panasonic Corporation, Osaka, Japan}
\begin{document}
\thispagestyle{fancy}
\pagenumbering{gobble}
\fancyhf{} 
\renewcommand{\headrulewidth}{0pt}
\setlength{\voffset}{-1.5cm}
\setlength{\headsep}{1.5cm}
\chead{Published as a conference paper at IEEE SLT 2016}
\cfoot{Copyright 2016 IEEE. Published in the 2016 IEEE Workshop on Spoken Language Technology (SLT 2016).}
%
\maketitle
\begin{abstract}
The fifth Dialog State Tracking Challenge (DSTC5) introduces a new cross-language dialog state tracking scenario, where the participants are asked to build their trackers based on the English training corpus, while evaluating them with the unlabeled Chinese corpus. Although the computer-generated translations for both English and Chinese corpus are provided in the dataset, these translations contain errors and careless use of them can easily hurt the performance of the built trackers. To address this problem, we propose a multichannel Convolutional Neural Networks (CNN) architecture, in which we treat English and Chinese language as different input channels of one single CNN model. In the evaluation of DSTC5, we found that such multichannel architecture can effectively improve the robustness against translation errors. Additionally, our method for DSTC5 is purely machine learning based and requires no prior knowledge about the target language. We consider this a desirable property for building a tracker in the cross-language context, as not every developer will be familiar with both languages.
\end{abstract}
\begin{keywords}
Convolutional neural networks, multichannel architecture, dialog state tracking, dialog systems
\end{keywords}
\section{Introduction}
\label{sec:intro}

Dialog state tracking is one of key sub-tasks of dialog management, whose goal is to transfer human utterances into a slot-value representation (dialog state) that is easy for computer to process and track the information that appeared in the dialog. To provide a common testbed for this task, the series of Dialog State Tracking Challenges (DSTC) was initiated \cite{DSTC5}. This challenge has already been held for four times, during which it provided a very valuable shared recourse for the research in this field and helped to improve the state-of-the-art. Since the forth challenge (DSTC4 2015), the target of dialog state tracking has been shifted from human-machine dialog to human-human dialog, which significantly increases the difficulty of the dialog state tracking task because of the variety and ambiguity in the human-human dialog. One lesson we learned from DSTC4 is the difficulty of building a high performance tracker for human-human dialog with very limited training corpus, no matter whether using machine learning or hand-crafted rule-based approaches \cite{CNN-DSTC4, dernoncourt2016robust}.　This is a very unfavorable situation because building hand-annotated training corpus is very expensive, time-consuming and requires human experts. Not to mention the collection of a new corpus for each language other than English if we need to build trackers for a new language.

The DSTC5 proposed a new challenge based on using the rapidly advancing machine translation (MT) technology, one may be able to adapt the built tracker to a new language with limited training or development corpus in that language. We find this idea very attractive because not only it can reduce the cost of new language adaptation, but also it provides the possibility of building a tracker with cross-language corpus. For example it can be very useful for developing the tourist information systems because one may have corpus collected from different language speakers (i.e. tourists from different countries): for each language, the amount of corpus may be very limited, but together it can be large enough for a good training. On the other hand, although the machine translation technology has achieved great progress recently, the translation quality is still not satisfactory \cite{sutskever2014sequence}. A conventional monolingual tracker trained on the computer-generated translations may lead to an imperfect model, and it can only accept the translations from other languages as input which will also degrade the performance.

To address these problems, we propose a model that can be trained with different languages at the same time, and use both original utterances and their translations as input source for the dialog state tracking. In such way, we can avoid building the tracker only based on computer-generated translations, and maximize the use of all possible input languages to increase the robustness to translation errors. This paper is organized as follows. Sect.\ref{sec:dataset} briefly describes the dataset and the dialog state tracking problem; Sect.\ref{sec:method} presents an overview of our method and explains in detail our multichannel CNN model. Sect.4 presents the evaluation results with analysis and discussion. Sect.5 concludes our work and proposes future improvements.

\section{DATASET AND PROBLEM DESCRIPTION}
\label{sec:dataset}

\begin{table}[t]
	\caption{Example of a transcription and dialog state label for a sub-dialog segment in the topic of `Accommodation'.}
	\label{example-sub-dialog}
	\begin{center}
	\resizebox{8.5cm}{!}{
	\begin{tabular}{p{240pt}|p{75pt}}
		\hline
		\centering{\bf Transcription} & {\bf \ Dialog state label} \\
		\hline
		{\bf Guide:} Let's try this one, okay? & \\
		{\bf Tourist:} Okay. & \\
		{\bf Guide:} It's InnCrowd Backpackers Hostel in Singapore. If you take a dorm bed per person only twenty dollars. If you & \\
		take a room, it's two single beds at fifty nine dollars & {\bf INFO:} Pricerange \\
		{\bf Tourist:} Um. Wow, that's good. & \\
		{\bf Guide:} Yah, the prices are based on per person per bed or dorm. But this one is room. So it should be fifty nine for the two room. So you're actually paying about ten dollars more per person only. & {\bf PLACE: } \hspace{30pt} InnCrowd Backpackers Hostel \\
		{\bf Tourist:} Oh okay. That's- the price is reasonable actually. It's good. & \\
		\hline
	\end{tabular}	
	}
	\end{center}
\end{table}

The fifth Dialog State Tracking Challenge (DSTC5) uses the whole dataset (including train/dev/test datasets) from the DSTC4 as the training dataset. This dataset contains 35 dialog sessions on tourist information for Singapore collected from English speakers. Besides the training dataset, a development set which includes 2 dialog sessions collected from Chinese speakers is provided for testing and tuning the trackers' cross-language performance before the final evaluation. Both the training and development sets are labelled with the dialog state tags and come with 5-best hypothesis of English or Chinese translations by a machine translation systems. In the evaluation phase of the challenge, a Test set including 8 unlabeled Chinese dialogs is distributed to each participant, and all prediction results submitted by each participant are evaluated by comparing with the true labels. The test dataset also includes 5-best English translations which are generated by the same machine translation system as the training/development dataset.

The dialog state in this challenge is defined by the same ontology used in DSTC4, which contains 5 topic branches with different slot sets (Table\ref{topic-slot}). These topic-slot combinations indicate the most important information mentioned in that topic, for example the `CUISINE' slot under the topic of `FOOD' refers to the cuisine types, while the `STATION' slot for the topic of `TRANSPORTATION' refers to the train stations. In total there are 30 such topic-slot combinations, and all possible values for each topic-slot are given as a list in the ontology. The main task of DSTC5 is to predict the proper value(s) for each slot given the current utterance and its topic, with all dialog history prior to that turn (e.g. Table\ref{example-sub-dialog}).

\begin{table}[h]
	\caption{List of slots for each topic.}
	\label{topic-slot}
	\begin{center}
	\resizebox{8.5cm}{!}{
	\begin{tabular}{c|p{250pt}}
		\hline
		{\bf Topic} & {\bf \hspace{100pt}SLOT} \\
		\hline
		{\bf Food} & INFO, CUISINE, TYPE\_OF\_PLACE, DRINK, PLACE, MEAL\_TIME, DISH, NEIGHBOURHOOD \\
		\hline
		{\bf Attraction} & INFO, TYPE\_OF\_PLACE, ACTIVITY, PLACE, TIME, NEIGHBOURHOOD \\
		\hline
		{\bf Shopping} & INFO, TYPE\_OF\_PLACE, PLACE, NEIGHBOURHOOD, TIME \\
		\hline
		{\bf Transportation} & INFO, FROM, TO, STATION, LINE, TYPE, TICKET \\
		\hline
		{\bf Accommodation} & INFO, TYPE\_OF\_PLACE, PLACE, NEIGHBOURHOOD \\
		\hline
	\end{tabular}	
	}
	\end{center}
\end{table}

\section{METHOD}
\label{sec:method}

In DSTC4 we proposed a method which is based on the convolutional neural networks originally proposed by Kim \cite{kim2014convolutional}. By this method we were able to achieve the best performance for tracking the “INFO” slot. The CNN model we used in this method was modified from the origin by adding a structure of multi-topic convolutional layer, so that it can better handle the information presented in different dialog topics. This model is characterized by its high performance for limited training data, because it can be trained across various topics. More details about this multi-topic model can be found in \cite{CNN-DSTC4}.

\begin{figure}[htb]
\includegraphics[width=8.5cm]{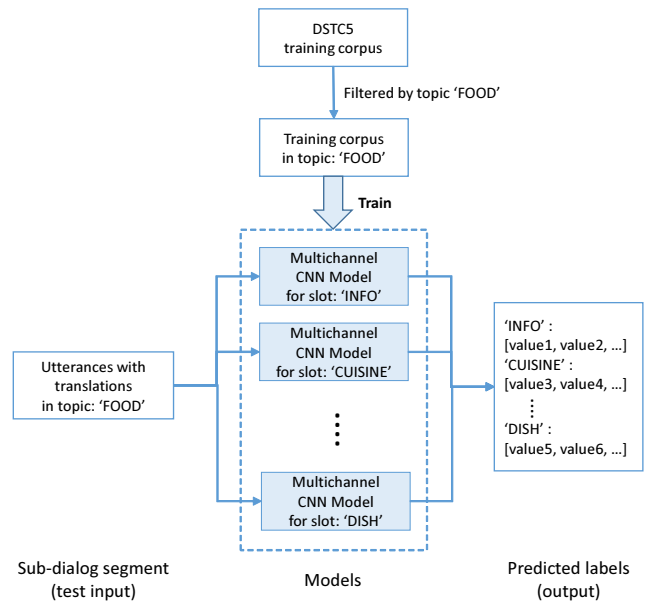}
\caption{Overview of our method (for the `FOOD' topic).}
\label{fig:method}
\end{figure}

In DSTC5 the training data is 75\% more than DSTC4, therefore the situation of limited training data is improved. In order to focus more on the new cross-language problem and keep our method simple, instead of using the more complex multi-topic model we proposed last time, we trained individual CNN model for each slot-topic combination. That is, for example the `INFO' slot in the topic of `FOOD' and the same `INFO' slot in the topic of `SHOPPING' will be trained by two independent models. This is the major difference from the last time, where we trained one single model for all topics. With this new scheme, we can set the hyperparameters in each model for each slot/topic to be exactly the same, so that our method is scalable, universally applicable and easy to tune. Fig \ref{fig:method} is a simple diagram illustrating our method.

\subsection{Motivation}
\label{ssec:motivation}

The biggest challenge of DSTC5 is that the training and test corpora are originally collected in different languages. Since both computer-generated Chinese and English translation are provided in the training and test dataset respectively, one straightforward approach is to train a model with English corpus and use it for the English translation in the test data. Alternatively, a model trained on Chinese translations in the training dataset can be used for Chinese utterances in the test data. However, both methods will waste the originally collected utterances either in the training or the test data. In order to fully utilize the corpus resource in both English and Chinese languages, we proposed the following multi-channel model which can be regarded as a combination of both English and Chinese models.

\subsection{Model architecture}
\label{ssec:model-architecture}

Our model is inspired by the multichannel convolutional neural networks commonly used in the image processing \cite{krizhevsky2012imagenet}. Instead of RGB channels used for color images, we apply each input channel to each different language source. In this model, the input of each channel is a two dimensional matrix, each row of which is the embedding vector of the corresponding word:
\begin{equation}
	\label{eq:segment-embedding}
	\mathbf{s} = \left[
	\begin{array}{c}
		\text{---}\ \mathbf{w}_{1}\ \text{---} \\
		\text{---}\ \mathbf{w}_{2}\ \text{---} \\
		\vdots\\
		\text{---}\ \mathbf{w}_{n}\ \text{---}
	\end{array}
	\right],
\end{equation}
where $\mathbf{w}_{i}\in\mathbb{R}^{k}$ is the embedding vector for the $i$-th word in the input text. This 2-dimensional array $\mathbf{s}$ is a matrix representation of the input text. We used three different word embedding in our model --- two for Chinese and one for English. The details of these embedding will be explained later in the Sect. \ref{ssec:embedding-models}. For each channel, a feature map $\mathbf{h}\in\mathbb{R}^{n-d+1}$ is obtained by convolving a trainable filter $\mathbf{m}\in\mathbb{R}^{d\times k}$ with the embedding matrix $\mathbf{s}\in\mathbb{R}^{n\times k}$ using the following equation:
\begin{equation}
	\mathbf{h} = f(\mathbf{m} \ast \mathbf{s} + \mathbf{b}).
\end{equation}
Here $f$ is a non-linear activation function\footnote{We used rectified linear unit (ReLU) for this activation function.}; $*$ is the convolution operator and $\mathbf{b} = (b,\ldots,b) \in\mathbb{R}^n$ is a bias term. The maximum value of this feature map $\hat{h} = \mathrm{max}\{\bm{h}\}$ is then selected by the max-pooling layer. This is the process how one filter extracts one most important feature from the input matrix. In this model, multiple filters are used in each channel to extract multiple features. These features then form the pooling layer which are passed to a fully connected layer for prediction.

The idea of multichannel model is to connect those extracted features from different channels before the final output, so that the model can use richer information obtained from different channels. The fully connect layer in multi-channel model follows the equation:
\begin{equation}
	\mathbf{y} = S\big(\mathbf{w}\cdot (\mathbf{\hat{h}}_{\mathrm{ch}1}\oplus \mathbf{\hat{h}}_{\mathrm{ch}2} \oplus \mathbf{\hat{h}}_{\mathrm{ch}3}) + \mathbf{b}\big),
\end{equation}
where $S$ is the sigmoid function; $\oplus$ is the concatenation operator and $\mathbf{\hat{h}}_{\mathrm{ch}n} = (\hat{h}_1,\ldots,\hat{h}_m)$ is the penultimate layer of the $n$-th channel.

\begin{figure}[!b]
\includegraphics[width=8.5cm]{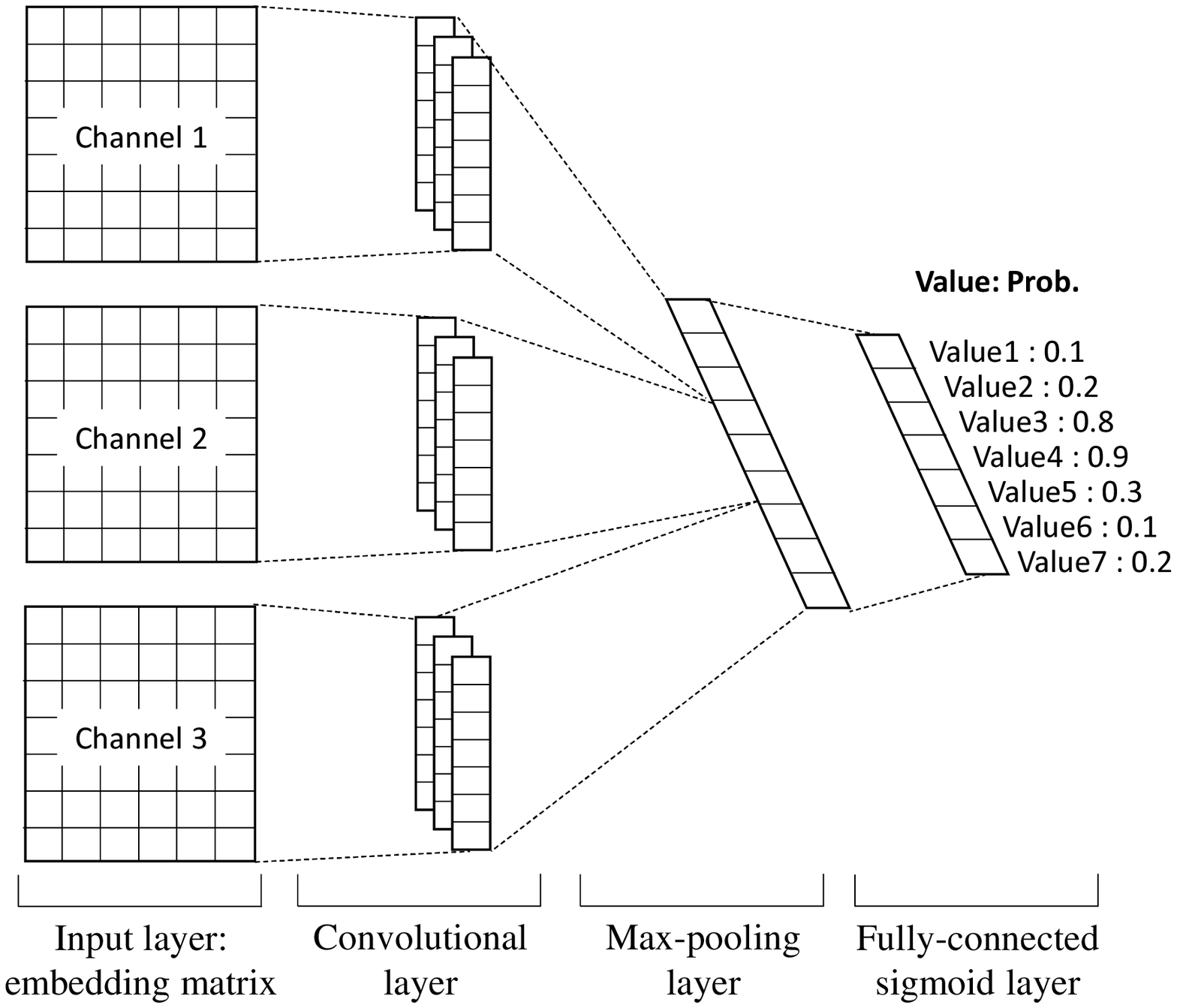}
\caption{Multichannel CNN model architecture for three input channels.}
\label{fig:model-architecutre}
\vspace{0.5cm}
\includegraphics[width=8.5cm]{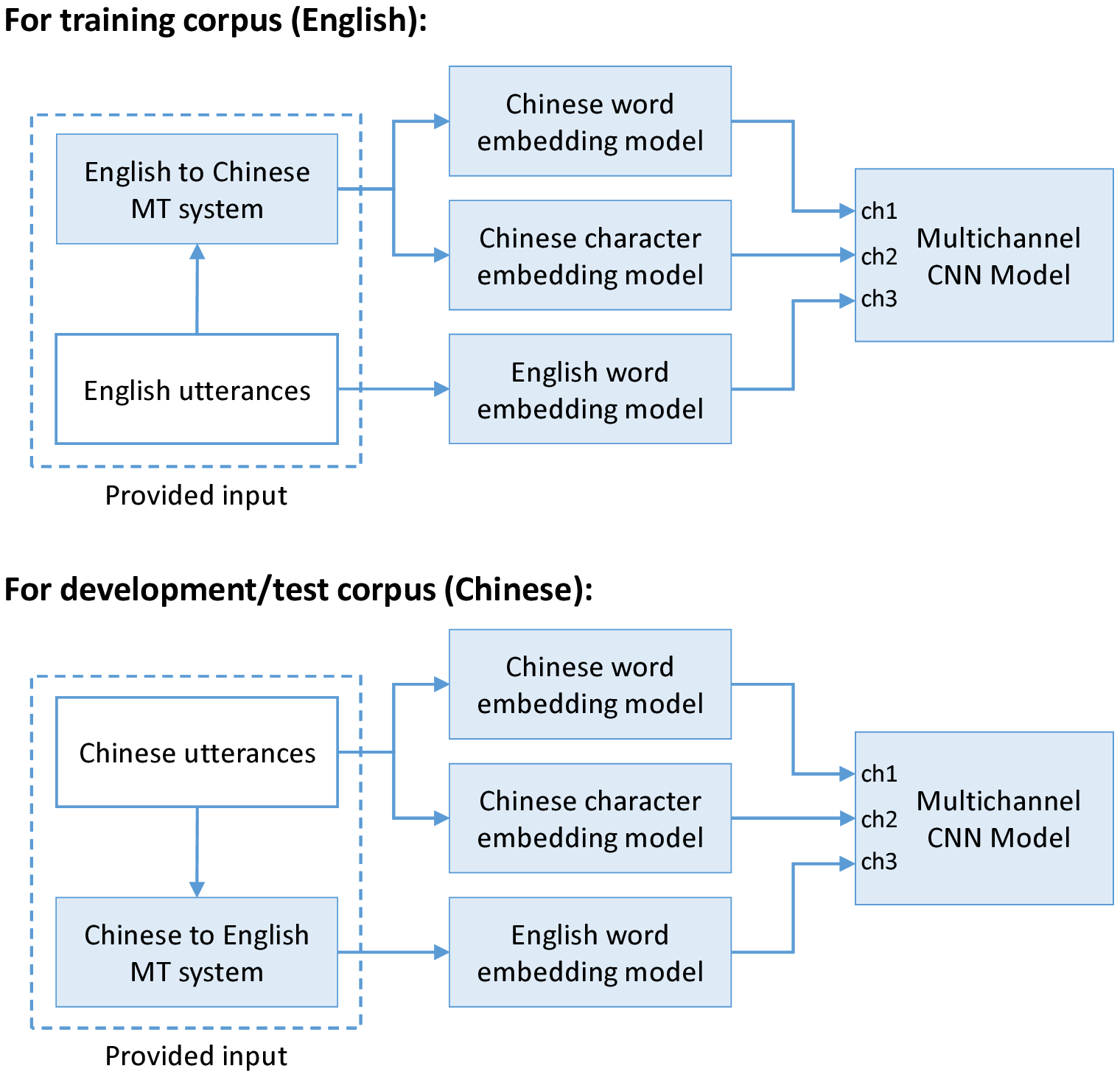}
\caption{Pre-processing of the input utterances.}
\label{fig:model-usage}
\end{figure}

Notice that in the original paper of Kim, a multi-channel architecture has also been proposed. The main difference between our model and their model is that we use different sets of filters for each channel, while in their model the same filter set are applied to all channels. The reason for this modification is that the word embedding for different languages can vary greatly, for example the same (or nearly the same) embedding vector in different language models may correspond to irrelevant words with very different meanings. Using different sets of filters ensures that proper features can be extracted in each channel no matter how the word embedding varies among different languages.

\subsection{Embedding models}
\label{ssec:embedding-models}
 
The word2vec \cite{mikolov2013efficient} is one of the most common methods for producing word embeddings. In DSTC5, we applied this method and trained three different models with different training corpus. The details of these models are listed as below: 
\begin{enumerate}
 \item English word model: 200-dimension word2vec model trained on English Wikipedia, with all text split by space and all letters lowercased. This model contains 253854 English words.
 \item Chinese word model: 200-dimension word2vec model trained on Chinese Wikipedia, with all text split by word boundary using `jieba' module\footnote{https://github.com/fxsjy/jieba}. This model contains 457806 Chinese words and 53743 English words appeared in the Chinese Wikipedia.
 \item Chinese character model: 200-dimension word2vec model trained on Chinese Wikipedia with all text split into single Chinese character. This model contains 12145 Chinese characters and 53743 English words appeared in the Chinese Wikipedia.
\end{enumerate}

The reason why we trained two models for Chinese language is because identifying word boundaries in Chinese is not a trivial task. For Chinese, the smallest element with meaning (word) varies from one single Chinese character to several concatenated Chinese characters, and the task for Chinese word splitting usually involves parsing the sentence and the state-of-the-art method still cannot achieve perfect accuracy. For this reason the Chinese word model may contain incorrect vocabularies and is not capable of handling unseen Chinese character combinations. On the other hand, the Chinese character model does not rely on word segmentation so that the model is error-free, and also it can easily deal with unseen words. However, since the Chinese character model ignores the word boundaries, the resulting embedding vector may not be able to reflect the precise meaning of each word.

\section{RESULTS}
\label{sec:results}

The results of the proposed method along with the scores of other teams are shown in the Table \ref{tab:results}. Our multichannel CNN model achieves the best score among all 9 teams: the result of entry-3 outperforms the second best team by 50\% (0.0956/0.0635) in Accuracy and 15\% (0.4519/0.3945) in F-measure with the sub-dialog evaluation. Our submitted five entries are the results of 5 different hyperparameters settings which are determined by a rough grid search\footnote{A guide for setting these hyperparameters can be found in \cite{zhang2015sensitivity}}, and those settings are summarized in Table \ref{tab:results-parameters}. Compared these results with each other, one can easily tell that among these hyperparameters the dropout rate is a key factor. The dropout is known as a technique for reducing overfitting in neural networks \cite{hinton2012improving}, and in our case reducing the dropout rate always improves the Precision while degrading the Recall score. One explanation for this is that an over-fitted model only outputs the same labels for the data which are very similar to the training data, and therefore decreases its generalization to unseen data. On the other hand, further decreasing the dropout rate does not improve the overall performance, whose results and parameter settings are also shown in the table as `Additional Expt. \#5\&6'.

\begin{table}[h]
	\caption[Caption for LOF]{Evaluation results (subdialog-level) on DSTC5 test dataset.\setcounter{footnote}{3}\footnotemark}
	\label{tab:results}
	\begin{center}
	\resizebox{8.5cm}{!}{
	\begin{tabular}{c|p{1.5cm}p{1.5cm}p{1.5cm}p{1.5cm}}
		\hline
		{\bf Tracker} & \centering{\bf Accuracy} & \centering{\bf Precision} & \centering{\bf Recall} & \multicolumn{1}{c}{\bf F-measure}\\
		\hline
		Multichannel \#3 & \centering{0.0956} & \centering{\bf 0.5643} & \centering{0.3769} & \multicolumn{1}{c}{\bf 0.4519}\\
		Multichannel \#4 & \centering{0.0872} & \centering{0.5427} & \centering{0.3842} & \multicolumn{1}{c}{0.4499}\\
		Multichannel \#0 & \centering{\bf 0.0964} & \centering{0.5217} & \centering{0.3849} & \multicolumn{1}{c}{0.4430}\\
		Multichannel \#1 & \centering{0.0712} & \centering{0.4340} & \centering{0.4196} & \multicolumn{1}{c}{0.4267}\\
		Multichannel \#2 & \centering{0.0681} & \centering{0.4216} & \centering{\bf 0.4303} & \multicolumn{1}{c}{0.4259}\\
		\hline
		Team4-entry2 & \centering{0.0635} & \centering{0.3768} & \centering{0.4140} & \multicolumn{1}{c}{0.3945}\\
		Team1-entry4 & \centering{0.0612} & \centering{0.3811} & \centering{0.3548} & \multicolumn{1}{c}{0.3675}\\
		Team6-entry1 & \centering{0.0383} & \centering{0.4063} & \centering{0.3124} & \multicolumn{1}{c}{0.3532}\\
		Team5-entry0 & \centering{0.0520} & \centering{0.3637} & \centering{0.3044} & \multicolumn{1}{c}{0.3314}\\
		baseline 2 & \centering{0.0222} & \centering{0.1979} & \centering{0.1774} & \multicolumn{1}{c}{0.1871}\\
		\hline
		\hline
		Additional Expt \#5 & \centering{0.0949} & \centering{0.5786} & \centering{0.3689} & \multicolumn{1}{c}{0.4505}\\
		Additional Expt \#6 & \centering{0.0888} & \centering{0.5677} & \centering{0.3712} & \multicolumn{1}{c}{0.4489}\\
		\hline
	\end{tabular}	
	}
	\end{center}
\end{table}
\footnotetext{These are the evaluation results using `Schedule 2' described in the challenge handbook \cite{DSTC5}.}

\begin{table}[h]
	\caption{Main hyperparameter settings for each entry.}
	\label{tab:results-parameters}
	\begin{center}
	\resizebox{8.5cm}{!}{
	\begin{tabular}{m{3.5cm}|ccccccc}
		\hline
		Entry \# in Table\ref{tab:results} & \#3 & \#4 & \#0 & \#1 & \#2 & \#5 & \#6 \\
		\hline\hline
		Dropout rate & 0.4 & 0.5 & 0.6 & 0.8 & 0.8 & 0.2 & 0 \\
		\hline
		Number of filters $\mathbf{m}\in\mathbb{R}^{1\times k}$ for each channel & 1000 & 600 & 1000 & 1400 & 1000 & 1000 & 1000 \\
		\hline
		Number of filters $\mathbf{m}\in\mathbb{R}^{2\times k}$ for each channel & 1000 & 600 & 1000 & 1400 & 1000 & 1000 & 1000 \\
		\hline
		{Learning rate} & \multicolumn{7}{c}{0.001} \\
		\hline
		{Weight decay coefficient for $L2$ regularization} & \multicolumn{7}{c}{0.0005} \\
		\hline
		{Training data} & \multicolumn{7}{c}{Training corpus + 1-best Chinese translation} \\
		& \multicolumn{7}{c}{\& Development corpus + 1-best English translation} \\
		\hline
		{Training epochs} & \multicolumn{7}{c}{100} \\
		\hline
		{Test input} & \multicolumn{7}{c}{Test corpus + 1-best English translation} \\
		\hline
	\end{tabular}	
	}
	\end{center}
\end{table}

\begin{table*}[t]
	\caption{Example of predicted labels by different models.}
	\label{tab:compare-example}
	\begin{center}
	\includegraphics[width=17cm]{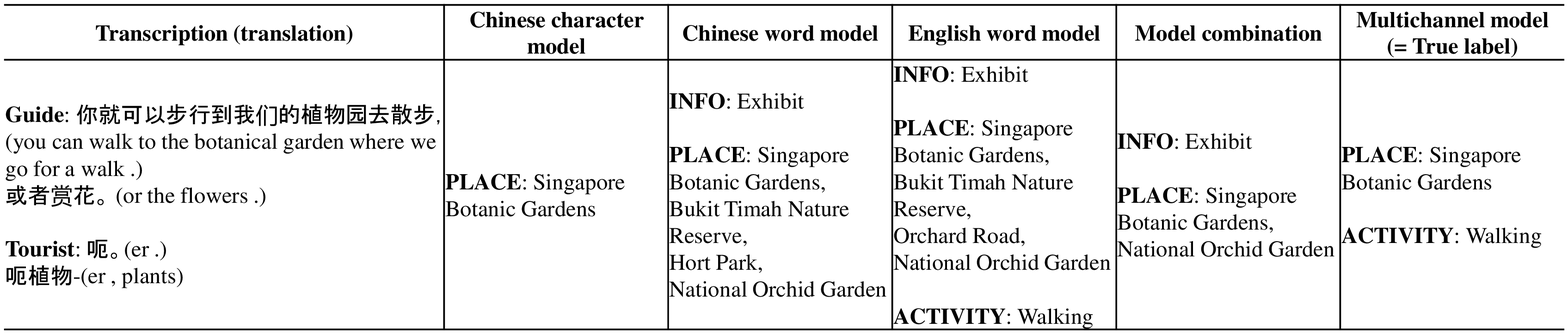}
	\end{center}
\end{table*}

\subsection{Multichannel model \& single channel model \& model combination}
\label{ssec:model-compare}

\begin{figure}[h]
\includegraphics[width=8.5cm]{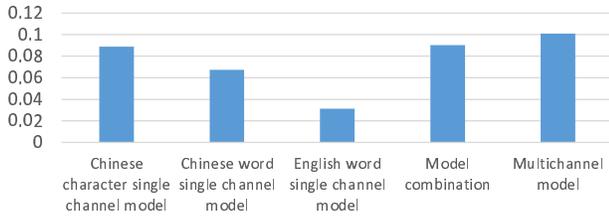}
\caption{Overall predictive accuracy of different models.}
\label{fig:model-compare}
\end{figure}

To investigate by how much the proposed multichannel architecture contributes to these results, we compared the performance between the multichannel and ordinary single channel CNN models. For this comparison, we trained three different monolingual single channel CNN models using each of the embedding models mentioned in Section\ref{ssec:embedding-models}. These  models used the same parameter setting as `multichannel \#3' in the Table \ref{tab:results-parameters}, and were trained only on the 1-best machine translation results. Fig.\ref{fig:model-compare} shows the comparing results: the Chinese character model achieves the best overall accuracy among single channel models, while the multichannel model outperforms all three single channel models.

In the earlier DSTC, a simple model combination technique has been used to further improve the predictive performance, where the final output is computed by averaging the scores output by different models \cite{henderson2014word}. We also applied this method to combine the output from all three single channel models, and the result is also shown in the Fig.\ref{fig:model-compare}. This simple model combination method does not perform as good as the multichannel model, but considering its simplicity, we still consider it as a good alternative to improve the performance.

\subsection{Discussion}
\label{ssec:discussion}

We think the above results can be partially explained from the point of view of ensemble learning. In a multichannel model, each channel provides a different view of the data, and an example is described using different feature sets that provide different, complementary information about the instance. The fully connected layer in the multichannel model further provides an optimization to use this information for the prediction, and therefore the resulting model can in principle better deal with the translation errors appeared in different channels. 

Table \ref{tab:compare-example} is one of the examples that demonstrate this idea. In this particular sub-dialog segment, none of the 3 single channel models is able to output the correct labels, while the multichannel model gives the correct prediction. As seen in this example, the model combination behaves like a simple voting, which means it only picks up the labels that are supported by majority of the single channel models. The multichannel model, on the other hand, is able to selectively choose which language source to trust more for each particular slot or value. As a result, the label of `Walking' is correctly predicted despite it only appearing in the English model's output, while the `Exhibit' label is correctly rejected even though it is supported by two single channel models out of three. 

\begin{figure}[!b]
\includegraphics[width=8.5cm]{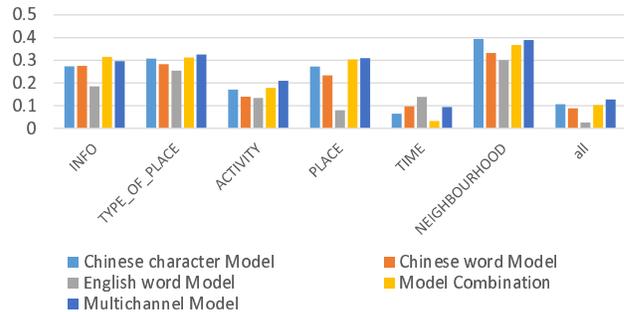}
\caption{Comparison of accuracy for each slot in the topic of `Attraction'.}
\label{fig:model-compare2}
\end{figure}

However the real situation is more complex. When we look at the overall predictive accuracy for each slot (Fig.\ref{fig:model-compare2}), we can find that the performance for each model varies on slots. We consider this is due to the ambiguity caused by machine translation which varies on different subjects. For example, as a time expression in English, 96\% of the word ``evening'' and 43\% of the word ``night'' are translated into the same Chinese word ``wan shang''. Although this Chinese word does have both meanings of ``evening'' and ``night'' in Chinese, there are more precise Chinese terms representing each word. This English to Chinese translation ambiguity immediately increases the difficulty of identifying the values of EVENING and NIGHT in Chinese, which leads to the poor performance of the Chinese model in the slot of TIME.

Another problem is that the translation quality often varies by reversing the translation direction, due to the difference in inflections, word order and grammars \cite{vilar2006error}. Since the training corpus only contains one translation direction (English to Chinese), the multichannel model is by no means optimized for the reverse translation direction. This may cause the multichannel model to have bias on certain channels, and it can explain why in certain slots the model combination that treats each channel equally works better. A more sophisticated way to train our multichannel model should be firstly training the model with one translation direction and then fine-tuning the model with the other. Unfortunately this is difficult in DSTC5 because the development dataset that can be used for the fine tuning is too limited.

\section{CONCLUSION}
\label{sec:conclusion}

We proposed a multichannel convolutional neural network, in which we treat multiple languages as different input channels. This multichannel model is found to be robust against the translation errors and outperforms any of the single channel models. Furthermore, our method does not require prior knowledge about new languages, and therefore can be easily applied to available corpus resources of different languages. This not only can reduce the cost for the adaption to a new language, but also offers the possibility to build multilingual dialog state trackers with large-scale cross-language corpora.

In this work we applied three different embedding models, while there is one more we did not try --- the English character model. There are several character-aware language models proposed recently, which are superior in dealing with subword information, rare words and misspelling \cite{kim2015character, zhang2015character}. We believe that integrating them into the multichannel model is a promising research direction. 

On the other hand, since our method is purely machine learning based, it cannot handle unseen labels in the test data. This is a very important issue especially for a large ontology, because of the difficulty in obtaining large training corpus that covers all concepts. To overcome this disadvantage, future work should include combining machine learning with other approaches, such as hand-craft rules, data argumentation and so on.

\bibliographystyle{IEEEbib}
\bibliography{strings,refs}

\end{document}